\documentclass[final]{nesy2026} 

\usepackage{longtable}
\usepackage[frozencache,cachedir=_minted]{minted}
\usepackage{float}

\usepackage{booktabs}
\usepackage[load-configurations=version-1]{siunitx} 
\usepackage{tcolorbox}
\tcbuselibrary{skins, breakable}

\theorembodyfont{\upshape}
\theoremheaderfont{\scshape}
\theorempostheader{:}
\theoremsep{\newline}

\title[NeSy-RAG]{NeSy-RAG: Neuro-Symbolic RAG for \\ Explainable Question Answering}

  \clearauthor{\Name{Jonas Gann} \Email{gann@informatik.uni-heidelberg.de}\\
   \Name{Michael Gertz} \Email{gertz@informatik.uni-heidelberg.de}\\
   \addr Data Science Group, Heidelberg University, Germany}

\begin{document}

\maketitle
\begin{abstract}
Retrieval-augmented generation (RAG) improves question answering by grounding large language models (LLMs) in external knowledge such as text corpora. However, its reasoning process remains largely opaque: intermediate reasoning steps are difficult to verify and cannot be reliably attributed to specific evidence. Moreover, missing user-specific context is rarely detected systematically, often leading to incomplete or incorrect output.

We propose \textbf{NeSy-RAG}, a modular neuro-symbolic RAG framework that synthesizes attributable Prolog modules from retrieved text chunks. For each chunk, the system generates semantically meaningful predicates that encode Boolean claims, which may depend on user facts. Using joint natural language--code embeddings, predicates are retrieved and composed into Prolog queries.
To address incomplete user context, we introduce a symbolic knowledge-gap detection mechanism that identifies missing user facts whose truth values affect the query outcome and automatically triggers follow-up interactions.

Executing the resulting Prolog queries yields deterministic answers together with transparent execution traces that link each reasoning step to its originating source. On the ShARC benchmark, without domain-specific training, NeSy-RAG achieves 61.1\% accuracy, outperforming a same-model RAG baseline that achieves 42.8\% accuracy.

\end{abstract}

\section{Introduction}
\label{sec:intro}
RAG combines LLMs with external knowledge and has become a standard approach to domain-specific question answering. By grounding answer generation in retrieved 
documents, RAG systems improve factual accuracy and reduce hallucinations \citep{Shuster2021RetrievalAR}. However, current RAG pipelines lack transparent and 
accountable reasoning: standard chain-of-thought (CoT) does not provide trustworthy explanations for LLM reasoning \citep{Chen2025ReasoningMD}, and the attribution of 
generated answers to source text is not guaranteed \citep{Zhou2024TrustworthinessIR}. In high-stakes domains such as healthcare or law, these limitations hinder trust and 
safe deployment.
Symbolic reasoning systems offer complementary strengths, including deterministic inference, interpretable reasoning traces, and explicit knowledge representations. 
Recent neuro-symbolic approaches leverage LLMs to generate or interact with symbolic programs, but they often rely on monolithic knowledge bases \citep{Zhang2025PrologRAGAS}, 
manually curated rules \citep{Tan2025PrologDrivenRD}, or dynamically generated logic programs without clear source attribution \citep{Borazjanizadeh2024ReliableRB, Di2025LoRPLL}.

\newpage

We propose \textbf{NeSy-RAG}, a neuro-symbolic RAG framework that addresses these limitations through novel Prolog synthesis methods. The main contributions are as follows:
\begin{itemize}
    \item A modular neuro-symbolic RAG framework that synthesizes and executes attributable Prolog modules from retrieved text chunks for explainable question answering.
    \item We introduce \emph{0-arity predicate abstraction}, which surfaces Prolog modules as Boolean claims, enabling constrained and scalable query construction via natural language to code retrieval.
    \item We propose a symbolic \emph{knowledge-gap detection} mechanism that enables the system to recognize and interactively resolve missing user information via follow-up questions.
    \item We empirically show that NeSy-RAG outperforms domain-agnostic LLM baselines on the ShARC \citep{Saeidi2018InterpretationON} benchmark.
\end{itemize}

\section{Related Work}

Recent work has explored integrating symbolic reasoning with LLMs through code synthesis. \citet{pan2023logic} propose a unified framework spanning logic programming, first-order logic, and constraint satisfaction problems, each realizable by different programming languages. Particularly Prolog's declarative nature makes it well-suited for LLM-based synthesis, enabling explainable and reliable reasoning with predicate logic.\newline
\textbf{Fact and Rule Attribution:} \citet{Yang2025NeuroSymbolicIB} highlight the explainability potential of Prolog by using execution traces as human-interpretable justifications for system outputs, enhancing trustworthiness and transparency. However, their approach does not address the attribution of rules and facts to the source knowledge itself. We extend this by ensuring that every fact and rule can be uniquely mapped to the text chunk from which it was derived.\newline
\textbf{Complex Context Information:} LLM-driven Prolog synthesis has been demonstrated for question answering \citep{Di2025LoRPLL,Borazjanizadeh2024ReliableRB}. However, reasoning over complex domains remains challenging, as extracting information for symbolic reasoning becomes increasingly difficult. \citet{Tan2025PrologDrivenRD} address this by incorporating domain knowledge to generate Prolog rules for medical diagnostics, but their approach relies on manually crafted specifications. Our method operates directly on complex, unstructured domain knowledge without requiring manual rule selection.\newline
\textbf{Knowledge Gaps:} While prior work on Prolog synthesis assumes that all information required for inference is available upfront, existing approaches do not address the challenge of missing user context at inference time. In practice, user-specific information is often required to provide accurate and personalized responses. \citet{li2024mediq} show that LLMs struggle to proactively seek missing information. We address this by proposing automated knowledge-gap detection that identifies missing user information during symbolic inference and proactively triggers targeted clarifying questions.\newline
\textbf{Large Domain Knowledge:} Dense retrieval is a well-established mechanism for aggregating relevant information across large corpora. \citet{Zhang2025PrologRAGAS} propose replacing dense retrieval with Prolog-based retrieval, but this can result in very large monolithic knowledge bases that are difficult to maintain and keep consistent. Our modular approach supports symbolic reasoning over large-scale domain knowledge without requiring a single unified knowledge base, enabling selective reasoning over relevant subsets while preserving scalability and symbolic integrity.\newline
\textbf{Integration:} \citet{Vakharia2024ProSLMAP} propose ProSLM, a Prolog-based question answering architecture that uses LLMs to translate natural language between the user and the Prolog system, laying the foundation for LLM-assisted Prolog question answering. However, as knowledge bases grow in size and complexity, formulating suitable Prolog queries that match both user intent and available symbolic knowledge becomes increasingly difficult. We address this bottleneck with a hierarchical predicate retrieval strategy that constrains query generation to a small, question-relevant set of 0-arity predicates, improving both scalability and query precision.

\section{NeSy-RAG Framework}
\label{section:methods}

\begin{figure}
    \centering
    \includegraphics[width=1\linewidth]{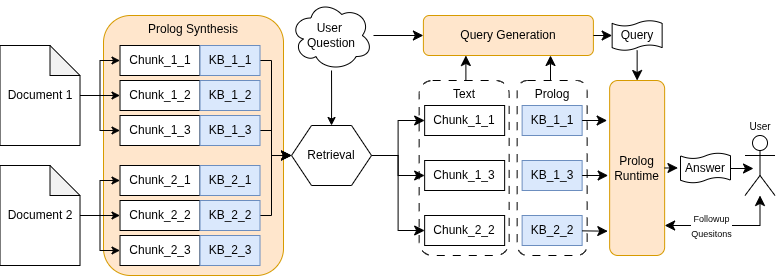}
    \caption{Overview of the proposed NeSy-RAG framework}
    \label{fig:architecture_overview}
\end{figure}

NeSy-RAG is designed as a plug-in extension to existing RAG systems, which are already widely adopted in practice. We consider a question answering setting in which a RAG system operates over domain knowledge $\mathcal{D} = \{d^{(1)}, \dots, d^{(M)}\}$ consisting of $M$ text documents. Each document $d^{(i)}$ is segmented into chunks by a chunking algorithm,
\(
chunk(d^{(i)}) = (c_1^{(i)}, \dots, c^{(i)}_{N_i}),
\)
and we denote the complete set of chunks as $\mathcal{C} = \bigcup_{i=1}^M chunk(d^{(i)})$.
Given a user question $q$, an existing retrieval mechanism $ret(q, \mathcal{C}) \subseteq \mathcal{C}$ aims to select a minimal subset of chunks sufficient for answering $q$. In practice, however, this subset may be incomplete, contain irrelevant chunks, or miss relevant ones entirely. 

NeSy-RAG takes these retrieved chunks as well as the question as input and proceeds with the symbolic reasoning pipeline described in the following sections. The overall performance of NeSy-RAG is therefore contingent on the quality of the underlying retrieval mechanism and domain knowledge. See Appendix \ref{apd:prompts} for LLM prompts used in this work.

The question $q$ is assumed to be \emph{Boolean} (e.g., ``Am I eligible for benefits?'') and be accompanied by a set of personal contextual information
\(
\mathcal{I} = \{i_1, \dots, i_j\},
\)
comprising user-specific attributes (e.g., ``I am a refugee.'', ``I am of legal age.'', ``I do not have UK citizenship.''). The contextual information $\mathcal{I}$ can be incomplete, making targeted follow-up questions to the user necessary to answer the question.

\subsection{Attributable Prolog Synthesis}
\label{subsection:attributable-prolog-synthesis}

As shown in Figure \ref{fig:architecture_overview}, for each retrieved text chunk \( c_m \in ret(q, \mathcal{C}) \), a Prolog module $p_m$ is synthesized. The synthesis itself can be performed by an LLM as proposed by \citet{Di2025LoRPLL} and \citet{Borazjanizadeh2024ReliableRB} by providing the content of the source chunk to the LLM. Each Prolog module \( p_m \) consists of rules and facts
\(
p_m = ( \mathcal{R}_m, \mathcal{F}_m ),
\)
where \( \mathcal{R}_m \) and \( \mathcal{F}_m \) denote the rules and facts derived from the text chunk \( c_m \), respectively.
This approach reduces the complexity of the synthesis task as each chunk contains only a limited amount of information.
Additionally, the modular approach allows for unique attribution of rules and facts to specific text chunks. For a given rule $r_m \in \mathcal{R}_m$ or fact $f_m \in \mathcal{F}_m$, the chunk $c_m$ can be determined as a unique source.
As part of the Prolog module system, reasoning traces not only track facts and rules themselves but also the module (and hence the source chunk) from which they originated. This enables granular explainability and verifiability of the reasoning process, as each accessed fact and rule can reference the source text.
Another advantage of this approach is that by modularizing the Prolog knowledge base, modules can be attached to their respective text chunks during retrieval. This enables not only the aggregation of chunks based on a question but also the aggregation of relevant symbolic knowledge. This makes a true scalable utilization of symbolic knowledge possible.

\subsection{Knowledge-Gap Detection}
\label{subsection:knowledge-gap-detection}

Prolog facts can be categorized into \emph{universal facts} and \emph{dynamic facts}. The distinguishing characteristic between these two categories is that dynamic facts contain at least one argument whose value cannot be determined at synthesis time (e.g., "User is of legal age.") and must instead be instantiated during inference, whereas all arguments of universal facts are fully specified by the domain knowledge (e.g., "Age threshold to be of legal age is 18.").

Explicitly separating universal and dynamic facts during Prolog synthesis enables the integration of external software components that can acquire or validate missing values for dynamic facts at inference time. Let a fact be defined as
\(
f_i = (h_i, A_i),
\)
where \(h_i\) denotes the predicate symbol (e.g., \texttt{of\_legal\_age}) and \(A_i\) are the arguments of the predicate. Each argument \(a_j \in A_i\) is represented as
\(
a_j = (n, t, v),
\)
where \(n\) denotes the argument name (e.g., \texttt{truth\_value}), \(t\) denotes a datatype (\texttt{bool}, \texttt{string}, \texttt{integer} or \texttt{float}), and \(v\) denotes a value of type \(t\). A fact is classified as a \emph{dynamic fact} if, for at least one of its arguments $a_j \in A_i$, the value cannot be inferred from domain knowledge.
For simplicity, we focus on dynamic facts with one Boolean argument, as they are sufficient for many question answering tasks and allow for straightforward user interactions (e.g., \texttt{of\_legal\_age(truth\_value: bool = true)}). However, the proposed approach can be extended to handle more complex data types and argument structures.

Given a dynamic fact \(df_i\) with a single Boolean argument \(a_1\), an interaction method \(im_i\) can be generated to acquire or validate the value of \(a_1\) during inference. If the value \(v\) of \(a_1\) is not specified in the Prolog fact, the user is asked to provide a value of the appropriate datatype. If a value \(v\) is already specified as part of a rule condition, the user is asked to confirm or reject it.
For each dynamic fact \(df_i\), its corresponding interaction method \(im_i\) is called whenever the Prolog system evaluates the fact. The values acquired in \(im_i\) are asserted in Prolog and the Boolean return value of the function is forwarded as the result of the dynamic fact.

\subsection{Question Answering}

After Prolog modules have been generated, they are used to answer questions as part of the inference pipeline.
As shown in Figure \ref{fig:architecture_overview}, given a user question, relevant text chunks are retrieved based on their vector embeddings using a retrieval component. In addition to text chunks, the corresponding Prolog knowledge bases can be attached as metadata. To answer a natural language question of a user given the retrieved knowledge bases, a Prolog query must be constructed that matches the user question using available predicates. After executing the query on the available knowledge bases, a Prolog response as well as an execution trace are obtained. The Prolog response can be further processed to provide an easy to understand answer for the user in natural language. The following sections go into more detail on each of the important subtasks of the question-answering pipeline.

\subsubsection{Query Translation}
\label{subsubsection:query-generation}

Given retrieved chunks and their Prolog modules, NeSy-RAG must construct a Prolog query that matches the user question while using only the available predicates.
Existing work translates questions into Prolog by providing an LLM the complete knowledge base \citep{Vakharia2024ProSLMAP}. However, this becomes brittle and unscalable as the knowledge base grows, since predicates of non-zero arity require the model to inspect the underlying knowledge base to determine valid variable bindings.
We instead formulate query construction as \emph{predicate selection and composition}, eliminating the need to load the knowledge base during query generation. To this end, the LLM additionally generates \emph{0-arity} rules during module synthesis, exposing the module’s functionality as self-contained Boolean claims:

\begin{minted}{prolog}
can_not_get_winter_fuel_payment :-
    lives_in_cyprus(true); lives_in_france(true);
    lives_in_gibraltar(true); lives_in_greece(true).
\end{minted}

At inference time, 0-arity rules are extracted from the modules and ranked by cosine similarity to the user question using a joint NL--PL embedding model \citep{Kryvosheieva2025EfficientCE} for \emph{natural language to code retrieval} (NL2C). An LLM is then instructed to compose a query \emph{only} from the most relevant subset of these rules. This reduces the query search space and mitigates hallucinations by constraining generation to predefined semantic units. Additionally, 0-arity rules abstract away from specific variable bindings, making it easier for the LLM to match user questions to relevant logic without needing detailed knowledge of rule implementations.
This two-stage hierarchical retrieval (first retrieving relevant chunks, then filtering relevant 0-arity rules within their corresponding modules) ensures that the query-generating LLM is exposed only to a small, question-relevant subset of the knowledge base. This is critical since query generation represents the primary scalability bottleneck: unlike module synthesis, which always operates on a single chunk in isolation, query generation must jointly handle all retrieved modules. The hierarchical retrieval thus enables scalability to both large domain corpora (snippet retrieval) and large modules (predicate retrieval).

\subsubsection{Answer Generation}
\citet{Vakharia2024ProSLMAP} propose to employ an LLM to generate a final user-friendly answer based on the Prolog result and the initial user question. Additionally, \citet{Yang2025NeuroSymbolicIB} propose to take advantage of the traces of Prolog executions as an explanation for the generated answers. In our approach, the Prolog trace additionally contains information that can uniquely map each rule and fact to a source chunk and the document it was extracted from. This enables further verification and attribution of reasoning steps.


\section{Experiments}

To evaluate the effectiveness of NeSy-RAG, we implemented a prototype and assessed it on the ShARC \citep{Saeidi2018InterpretationON} benchmark, a conversational question answering dataset built from legal text sourced from the UK government website. Each entry consists of a \textbf{snippet} (a passage from the website), a \textbf{question}, a \textbf{scenario} describing the user's situation, a \textbf{history} of prior follow-up exchanges, \textbf{evidence} capturing user-specific information extractable from the scenario, and a target \textbf{answer} (\texttt{yes}, \texttt{no}, \texttt{irrelevant} or \texttt{more}).

Rather than supplying all user information upfront, we consolidate \texttt{scenario}, \texttt{evidence}, and \texttt{history} into a single \texttt{context} field that is \textbf{withheld} from the system at the start of each interaction. This design simulates a setting in which the system must actively recover missing user information through follow-up questions. Whenever the system poses a follow-up question, a separate LLM (as part of the evaluation system) answers it by consulting the \texttt{context}, thereby simulating an ideal user. This setup deliberately decouples:

\begin{itemize}
    \item \textbf{Object of evaluation}: Reasoning process and follow-up question generation based on \texttt{snippet} and \texttt{question}
    \item \textbf{Simulated user}: LLM generated answers to follow-up questions based on \texttt{context}
\end{itemize}

Given a \texttt{snippet} and a \texttt{question}, the evaluated systems are expected to return \texttt{yes} or \texttt{no} once sufficient user information has been gathered through follow-up questions. If the simulated user is unable to answer necessary follow-up questions, the system should predict \texttt{more}, signaling that the answer remains undetermined. If the snippet is unrelated to the question, the system should predict \texttt{irrelevant}. \tableref{tab:sample-input-output} illustrates an example of system input and output for NeSy-RAG.

The ShARC dataset is provided with training, development, and test splits. Hence, this dataset is targeted at prediction algorithms that need to be trained for the given domain. Instead of training on this domain-specific data, the presented approach is domain-agnostic and does not utilize any prior knowledge.
We evaluate on the ShARC test split (8,276 entries). Consistent with the RAG setup described at the beginning of Section \ref{section:methods}, the system receives a candidate set of snippets and must predict an answer. To simulate a retrieval result without implementing a specific retrieval process, each target snippet is presented alongside randomly sampled distractor snippets from the test split. Note that the target snippet itself may be unrelated to the question, in which case the correct prediction is \texttt{irrelevant} regardless of the distractors. As each ShARC instance relies on a single relevant snippet, multi-hop retrieval is outside the scope of this evaluation but important future work.

\begin{table}[h!]
\floatconts
  {tab:sample-input-output}
  {\caption{Sample input, context and output of NeSy-RAG for a single ShARC instance.}}
  {\small
  \begin{tabular}{lp{3.2cm}p{6.3cm}}
  \toprule
  \textbf{Input}
    & Snippet
      & If you don't claim marriage allowance and you or your partner
        were born before 6 April 1935, you may be able to claim
        married couple's allowance. \\[6pt]
    & Question
      & Could I claim married couple's allowance? \\
  \midrule
  \multicolumn{3}{l}{%
    \textbf{Context}: \quad  
    \begin{tabular}[t]{ll}
      Do you claim marriage allowance? & $\rightarrow$ \textbf{No} \\
      Were you or your partner born before 6 April 1935? & $\rightarrow$ \textbf{Yes}
    \end{tabular}} \\
  \midrule
  \textbf{Output}
    & Prolog Module
      & See Appendix \ref{apd:sample-knowledge-base} \\
  \cmidrule{2-3}
    & Prolog Query
      & \texttt{?-could\_claim\_married\_couples\_allowance.} \\
  \cmidrule{2-3}
    & Dynamic Predicates \newline {\scriptsize(part of Prolog module)}
      & \texttt{claims\_marriage\_allowance} \newline
        \texttt{born\_before\_6\_april\_1935} \\
  \cmidrule{2-3}
    & Follow-up Questions \newline {\scriptsize(answered via context)}
      & \texttt{Claim marriage allowance?} $=$ \textbf{false} \newline
        \texttt{Born before 06.04.1935?} $=$ \textbf{true} \\
  \cmidrule{2-3}
    & System Answer & \textbf{Yes} \\
  \midrule
  \textbf{Gold} & Target Answer & Yes \\
  \bottomrule

  \end{tabular}}
\end{table}

\subsection{Prototype Implementation}

We implemented a prototype of NeSy-RAG following the components described in Section~\ref{section:methods}. For each dataset instance, the system receives a set of snippets consisting of the instance snippet and additional distractors. An LLM classifies each snippet as \texttt{relevant} or \texttt{irrelevant}. If no snippet is classified as relevant, the system predicts \texttt{irrelevant}.

For each relevant snippet, the system synthesizes a Prolog module and augments it with 0-arity rules. Prolog modules can be reused for future questions, thus reducing computation time. To construct a query, all 0-arity rules are embedded and ranked by cosine similarity to the embedded user question. The top-$k$ rules are passed to an LLM, which composes a Prolog query using only these rules. If the LLM returns \texttt{None}, indicating that no suitable query could be constructed, the system predicts \texttt{irrelevant}.

The generated Prolog modules are loaded into \texttt{swipl}\footnote{SWI-Prolog: \url{https://www.swi-prolog.org}} and the query is executed. Whenever a dynamic predicate is encountered during execution, a corresponding follow-up question is posed to the simulated user via a foreign Python function implemented using \texttt{PySwip}\footnote{PySwip: \url{https://pypi.org/project/pyswip}}. If the simulated user is unable to answer a follow-up question, the influence of the unresolved predicate on the query outcome is assessed by evaluating it under both Boolean assignments. If the outcome varies across assignments, the system predicts \texttt{more}. Otherwise, the consistent outcome is returned as \texttt{yes} or \texttt{no}. To avoid identical follow-up questions, the foreign function caches evaluated dynamic predicates and their answers for the ongoing conversation. This can be extended to storing retrieved information in a user profile.

\subsection{Baseline Implementations}
\label{subsection:baseline_implementations}

We compare NeSy-RAG against two LLM baselines that differ in how they access user information and what snippets they receive. Both baselines predict the four labels \texttt{yes}, \texttt{no}, \texttt{irrelevant}, or \texttt{more}. The structured LLM output functionality of OpenAI and Ollama\footnote{Ollama: \url{https://ollama.com/}} is used to enforce permissible prediction classes.

\textbf{LLM Upper Baseline (without follow-up questions):}
We prompt an LLM to directly predict the target label given the full instance information,
including the \emph{singular} target \texttt{snippet}, the \texttt{question}, and the \texttt{context}. Since the \texttt{context} contains all available information from the user, this baseline does not need to generate follow-up questions. Therefore, it serves as an approximate upper bound for domain-agnostic LLM performance.

\textbf{LLM RAG Baseline (with follow-up questions):}
This baseline mirrors NeSy-RAG's setting but replaces Prolog with pure LLM inference. It first selects relevant snippets from the synthetic retrieval set, then predicts an answer to the \texttt{question} using only those snippets. If the initial prediction is \texttt{more}, the system generates a list of follow-up questions which are answered using the available \texttt{context}. Then the final label is predicted again while having access to the newly answered follow-up questions.

\subsection{Evaluation Results}
\label{sec:eval}

For the evaluation of the approaches, \texttt{gpt-oss:20b} and \texttt{gpt-5-mini} were used 
for LLM calls, and the code embedding model \texttt{jina-code-embeddings-1.5b} 
\citep{Kryvosheieva2025EfficientCE} was used for embedding and retrieval of 0-arity 
predicates in NeSy-RAG. \texttt{gpt-oss:20b} was run on an \texttt{NVIDIA RTX 3090} 
using Ollama. The number of retrieved snippets (including the instance snippet) is 3. Since per-snippet relevance filtering is performed independently, scaling the number of snippets is not expected to affect accuracy and was not evaluated.

\tableref{tab:sharc-results} presents the evaluation results, with accuracy measured as the ratio of correct class predictions and the duration as the average time in seconds to process a single dataset instance. \citet{Luo2023ExplicitAA} (BiAE) represents the current state of the art on ShARC and serves as an upper bound for trained approaches. BiAE receives \texttt{scenario} and \texttt{history} directly as 
input and therefore does not rely on intermediate follow-up questions, whereas NeSy-RAG must recover equivalent user context interactively. 

NeSy-RAG outperforms both LLM RAG baselines with 61.1\% accuracy for $k=10$ retrieved 0-arity predicates. Smaller and larger values of $k$ result in a small drop in accuracy. Among the RAG baselines, scaling to a larger model yields substantial accuracy gains (42.8\% to 53.7\%). NeSy-RAG may 
similarly benefit from a larger model, which remains to be explored. Beyond accuracy, NeSy-RAG achieves a lower mean execution time of 7.4s, compared to 11.4s (\texttt{gpt-oss:20b}) for the RAG baseline.
This efficiency stems from three factors: (1) Prolog modules are synthesized once per snippet and reused across questions, eliminating redundant LLM reads of source text; (2) query construction operates only over a small set of retrieved 0-arity predicates rather than the full module, keeping context windows small; and (3) follow-up questions are triggered algorithmically during Prolog execution, avoiding the additional LLM call required by the RAG baseline to generate them. These properties also suggest additional token savings over standard LLM-based pipelines, though a systematic evaluation remains future work.
Taken together, these results demonstrate that NeSy-RAG can surpass standard LLM-based approaches in both accuracy and execution time while relying on symbolic 
reasoning. The LLM upper baselines score substantially higher, achieving up to 69.9\% accuracy, which establishes a ceiling for non-trained LLM-based 
question-answering on this dataset. BiAE further raises this ceiling to 77.9\%, 
reflecting the advantage of task-specific training.

\begin{table}[h!]
\floatconts
  {tab:sharc-results}
  {\caption{Evaluation results for the ShARC dataset. NeSy-RAG accuracy is reported for k=3 / k=10 / k=20 retrieved 0-arity predicates.}}
  {\begin{tabular}{llll}
  \toprule
  \bfseries Approach & \bfseries Model & \bfseries Accuracy (\%) &\bfseries Avg. Duration (s) \\
  \midrule
  LLM RAG Baseline & gpt-oss:20b & 42.8 & 11.4 \\ 
  LLM RAG Baseline & gpt-5-mini & 53.7 & 25.1\\ 
  \textbf{NeSy-RAG} & gpt-oss:20b & 60.0 / 61.1 / 60.4 & 7.0 / 7.4 / 7.3 \\  
  LLM Upper Baseline & gpt-oss:20b & 64.1 & 2.7\\ 
  LLM Upper Baseline & gpt-5-mini & 69.9 & 5.2\\ 
  BiAE (trained) & - & 77.9 & -\\
  \bottomrule
  \end{tabular}}
\end{table}

\figureref{fig:confusion_nesy_rag} shows the normalized confusion matrices for NeSy-RAG and both LLM baselines, all evaluated using \texttt{gpt-oss:20b}. 178 NeSy-RAG executions (2.2\%) resulted in errors caused by syntactically invalid Prolog modules or queries. The LLM baselines exhibit sporadic errors, caused by occasional failures of the structured output mechanism.

\begin{figure}[h!]
\floatconts
  {fig:confusion_nesy_rag}
  {\caption{Normalized Confusion Matrices of NeSy-RAG and LLM Baselines}}
  {\includegraphics[width=\linewidth]{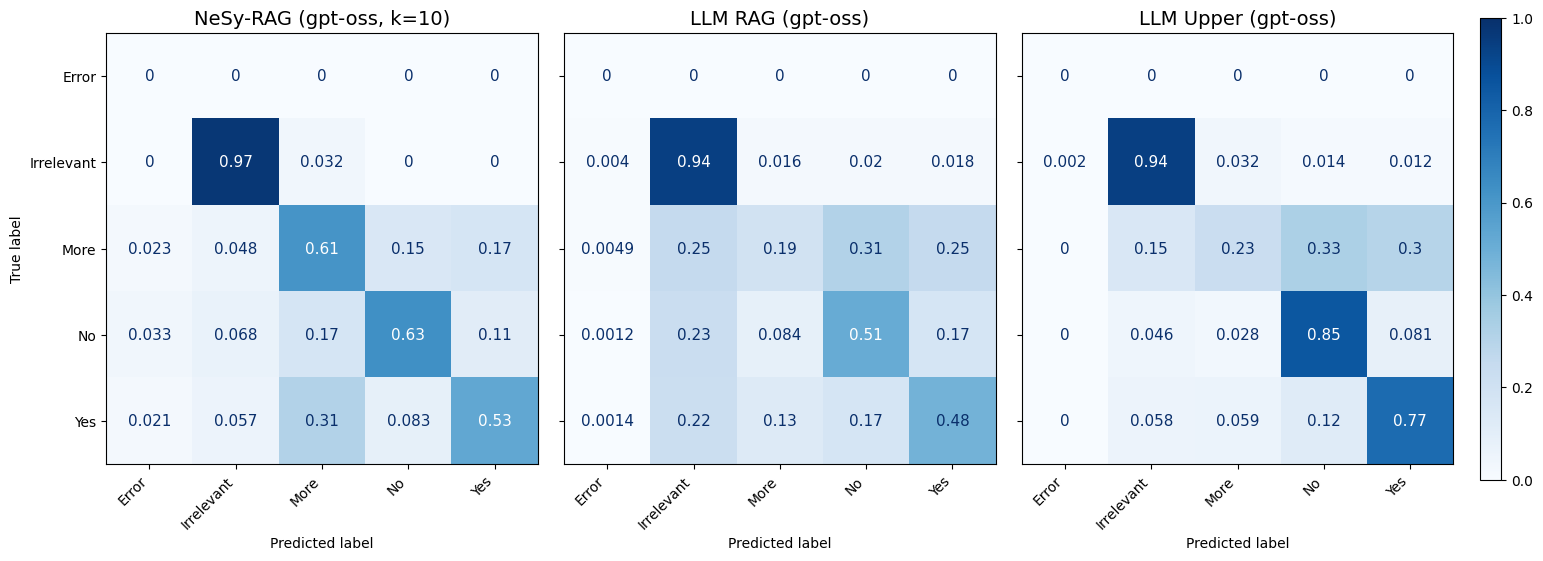}}
\end{figure}

NeSy-RAG correctly classifies 61\% of \texttt{more} instances, exceeding LLM RAG (19\%) and LLM Upper (23\%). This gap is strong empirical evidence for NeSy-RAG's symbolic knowledge-gap detection mechanism, a core novelty of this work. Both LLM baselines tend to commit to a definitive answer although essential user information is missing. On \texttt{yes} and \texttt{no}, however, LLM Upper correctly classifies a higher proportion of instances (77\% and 85\%) than NeSy-RAG (53\% and 63\%), reflecting the advantage of having complete user context available upfront. The LLM RAG baseline additionally exhibits a strong bias toward predicting \texttt{irrelevant} on answerable questions. This likely is because the model prematurely rejects snippets as irrelevant rather than generating follow-up questions to resolve uncertainty.

The dominant error pattern for NeSy-RAG is the misclassification of \texttt{yes} instances as \texttt{more} (31\%), suggesting that the synthesis step generates superfluous dynamic predicates that trigger unnecessary follow-up questions. Tightening dynamic predicate generation is therefore the most direct path to closing the remaining accuracy gap to LLM Upper. This conservative bias, however, trades accuracy for safety: in high-stakes domains, responding with \texttt{more} is preferable to false certainty. That said, NeSy-RAG still exhibits false certainty, misclassifying \texttt{more} instances as \texttt{yes} (17\%) or \texttt{no} (15\%), though both rates remain lower than those observed in either baseline.

\section{Discussion and Conclusion}

In this work, we presented NeSy-RAG, a novel modular neuro-symbolic RAG framework
that synthesizes and executes attributable Prolog modules from retrieved text chunks
for explainable question answering. By combining dense retrieval with symbolic
reasoning, NeSy-RAG directly addresses three core limitations of standard RAG
pipelines: opaque reasoning without source attribution, unscalable query construction
over large knowledge bases, and the inability to systematically detect missing user
information.

The modular Prolog synthesis approach ensures that every fact and rule can be uniquely
traced back to the text chunk from which it was derived, providing granular
explainability and verifiability that pure LLM-based pipelines cannot offer. The
0-arity predicate abstraction, combined with joint NL--PL retrieval, enables scalable
and constrained query construction: rather than exposing a query-generating LLM to an
entire knowledge base, NeSy-RAG performs hierarchical retrieval at both the chunk and
predicate level, keeping context windows small and mitigating hallucination. The knowledge-gap detection mechanism allows the system to algorithmically
identify missing user information during Prolog execution and trigger
follow-up interactions.

Our evaluation on the ShARC benchmark confirms the effectiveness of the proposed approach:
NeSy-RAG achieves 61.1\% accuracy, outperforming the domain-agnostic LLM RAG baseline
(42.8\%), while also achieving lower mean execution time owing to module reusability
and smaller context windows. NeSy-RAG correctly classifies 61\% of
\texttt{more} instances compared to 19\% for the LLM RAG baseline, providing evidence for the knowledge-gap detection mechanism. The primary remaining
limitation is over-generation of dynamic predicates, which causes some answerable
instances to be conservatively classified as \texttt{more}. In high-stakes domains, this conservative bias is arguably preferable to false certainty.

Several directions remain ongoing work. Tightening the constraints on dynamic
predicate generation is the most direct path to closing the remaining accuracy gap.
Extending NeSy-RAG to multi-hop and multi-chunk settings would further demonstrate the scalability of
modular Prolog synthesis and inter-module reasoning and attribution. Finally,
broadening knowledge-gap detection beyond Boolean predicates would increase
applicability to richer, more varied question answering settings.

\newpage

\bibliography{nesy2026-sample}

\begin{thebibliography}{14}
\providecommand{\natexlab}[1]{#1}
\providecommand{\url}[1]{\texttt{#1}}
\expandafter\ifx\csname urlstyle\endcsname\relax
  \providecommand{\doi}[1]{doi: #1}\else
  \providecommand{\doi}{doi: \begingroup \urlstyle{rm}\Url}\fi

\bibitem[Borazjanizadeh and Piantadosi(2024)]{Borazjanizadeh2024ReliableRB}
Nasim Borazjanizadeh and Steven Piantadosi.
\newblock {Reliable Reasoning Beyond Natural Language}.
\newblock \emph{ArXiv}, abs/2407.11373, 2024.

\bibitem[Chen et~al.(2025)Chen, Benton, Radhakrishnan, Uesato, Denison, Schulman, Somani, Hase, Wagner, Roger, Mikulik, Bowman, Leike, Kaplan, and Perez]{Chen2025ReasoningMD}
Yanda Chen, Joe Benton, Ansh Radhakrishnan, Jonathan Uesato, Carson~E. Denison, John Schulman, Arushi Somani, Peter Hase, Misha Wagner, Fabien Roger, Vladimir Mikulik, Sam Bowman, Jan Leike, Jared Kaplan, and Ethan Perez.
\newblock {Reasoning Models Don't Always Say What They Think}.
\newblock \emph{ArXiv}, abs/2505.05410, 2025.

\bibitem[Di et~al.(2025)Di, Zhang, Lv, Cui, and Liu]{Di2025LoRPLL}
Zhengkun Di, Chaoli Zhang, Hongtao Lv, Lizhen Cui, and Lei Liu.
\newblock {LoRP: LLM-based Logical Reasoning via Prolog}.
\newblock \emph{Knowl. Based Syst.}, 2025.

\bibitem[Kryvosheieva et~al.(2025)Kryvosheieva, Sturua, G{\"u}nther, Martens, and Xiao]{Kryvosheieva2025EfficientCE}
Daria Kryvosheieva, Saba Sturua, Michael G{\"u}nther, Scott Martens, and Han Xiao.
\newblock {Efficient Code Embeddings from Code Generation Models}.
\newblock \emph{ArXiv}, abs/2508.21290, 2025.

\bibitem[Li et~al.(2024)Li, Balachandran, Feng, Ilgen, Pierson, Koh, and Tsvetkov]{li2024mediq}
Shuyue~S Li, Vidhisha Balachandran, Shangbin Feng, Jonathan~S Ilgen, Emma Pierson, Pang~W Koh, and Yulia Tsvetkov.
\newblock {MediQ: Question-Asking LLMs and a Benchmark for Reliable Interactive Clinical Reasoning}.
\newblock \emph{Advances in Neural Information Processing Systems}, 37:\penalty0 28858--28888, 2024.

\bibitem[Luo et~al.(2023)Luo, Tian, Yuan, and Wang]{Luo2023ExplicitAA}
Yangyang Luo, Shiyu Tian, Caixia Yuan, and Xiaojie Wang.
\newblock {Explicit Alignment and Many-to-many Entailment Based Reasoning for Conversational Machine Reading}.
\newblock In \emph{Conference on Empirical Methods in Natural Language Processing}, 2023.

\bibitem[Pan et~al.(2023)Pan, Albalak, Wang, and Wang]{pan2023logic}
Liangming Pan, Alon Albalak, Xinyi Wang, and William Wang.
\newblock {Logic-LM: Empowering Large Language Models with Symbolic Solvers for Faithful Logical Reasoning}.
\newblock In \emph{Findings of the Association for Computational Linguistics: EMNLP 2023}, pages 3806--3824, 2023.

\bibitem[Saeidi et~al.(2018)Saeidi, Bartolo, Lewis, Singh, Rockt{\"a}schel, Sheldon, Bouchard, and Riedel]{Saeidi2018InterpretationON}
Marzieh Saeidi, Max Bartolo, Patrick Lewis, Sameer Singh, Tim Rockt{\"a}schel, Mike Sheldon, Guillaume Bouchard, and Sebastian Riedel.
\newblock {Interpretation of Natural Language Rules in Conversational Machine Reading}.
\newblock In \emph{Conference on Empirical Methods in Natural Language Processing}, 2018.

\bibitem[Shuster et~al.(2021)Shuster, Poff, Chen, Kiela, and Weston]{Shuster2021RetrievalAR}
Kurt Shuster, Spencer Poff, Moya Chen, Douwe Kiela, and Jason Weston.
\newblock {Retrieval Augmentation Reduces Hallucination in Conversation}.
\newblock In \emph{Conference on Empirical Methods in Natural Language Processing}, 2021.

\bibitem[Tan et~al.(2025)Tan, Li, Xu, Qu, Chu, Xu, Qi, and Qiu]{Tan2025PrologDrivenRD}
Xiaoyu Tan, Bin Li, Weidi Xu, Chao Qu, Wei Chu, Yinghui Xu, Yuan Qi, and Xihe Qiu.
\newblock {Prolog-Driven Rule-Based Diagnostics with Large Language Models for Precise Clinical Decision Support}.
\newblock In \emph{International Conference on Medical Image Computing and Computer-Assisted Intervention}, 2025.

\bibitem[Vakharia et~al.(2024)Vakharia, Kufeldt, Meyers, Lane, and Gilpin]{Vakharia2024ProSLMAP}
Priyesh Vakharia, Abigail Kufeldt, Max Meyers, Ian Lane, and Leilani~H. Gilpin.
\newblock {ProSLM: A Prolog Synergized Language Model for explainable Domain Specific Knowledge Based Question Answering}.
\newblock In \emph{International Workshop on Neural-Symbolic Learning and Reasoning}, 2024.

\bibitem[Yang et~al.(2025)Yang, Li, Cui, Bing, and Lam]{Yang2025NeuroSymbolicIB}
Sen Yang, Xin Li, Leyang Cui, Li~Bing, and Wai Lam.
\newblock {Neuro-Symbolic Integration Brings Causal and Reliable Reasoning Proofs}.
\newblock \emph{Findings of the Association for Computational Linguistics}, 2025.

\bibitem[Zhang et~al.(2025)Zhang, Li, Peng, Zheng, and Meng]{Zhang2025PrologRAGAS}
Bailing Zhang, Jiajie Li, Kang Peng, Shuchang Zheng, and Kai Meng.
\newblock {Prolog-RAG: A Symbolic Reasoning Approach to Retrieval-Augmented Generation}.
\newblock \emph{International Conference on Computer Engineering and Application}, 2025.

\bibitem[Zhou et~al.(2024)Zhou, Liu, Li, Jin, Qian, Liu, Li, Dou, Ho, and Yu]{Zhou2024TrustworthinessIR}
Yujia Zhou, Yan Liu, Xiaoxi Li, Jiajie Jin, Hongjin Qian, Zheng Liu, Chaozhuo Li, Zhicheng Dou, Tsung-Yi Ho, and Philip~S. Yu.
\newblock {Trustworthiness in Retrieval-Augmented Generation Systems: A Survey}.
\newblock \emph{ArXiv}, abs/2409.10102, 2024.

\end{thebibliography}

\newpage

\appendix

\section{Sample Prolog Knowledge Base}\label{apd:sample-knowledge-base}

The following is an example of a Prolog module that was created by NeSy-RAG for the text snippet shown in \tableref{tab:sample-input-output}.

\begin{minted}[fontsize=\small]{prolog}
:- dynamic claims_marriage_allowance/1.
% claims_marriage_allowance(Boolean) indicates whether the user or their
% partner currently claims Marriage Allowance (true) or does not (false).

:- dynamic born_before_6_april_1935/1.
% born_before_6_april_1935(Boolean) indicates whether the user
% or their partner was born before 6 April 1935 (true) or not (false).

% eligible_for_married_couples_allowance is true when the user is not claiming
% Marriage Allowance and either the user or their partner was born before 6 April 1935.
eligible_for_married_couples_allowance :-
    claims_marriage_allowance(false),
    born_before_6_april_1935(true).

% not_eligible_for_married_couples_allowance is true when the user is already claiming
% Marriage Allowance or neither the user nor their partner is born before 6 April 1935.
not_eligible_for_married_couples_allowance :-
    claims_marriage_allowance(true);
    born_before_6_april_1935(false).

% could_claim_married_couples_allowance is true when the user may be able to claim
% Married Couple’s Allowance because they are eligible.
could_claim_married_couples_allowance :-
    eligible_for_married_couples_allowance.

% can_not_claim_married_couples_allowance is true when the user is not eligible
% to claim Married Couple’s Allowance.
can_not_claim_married_couples_allowance :-
    not_eligible_for_married_couples_allowance.

% can_answer_married_couples_allowance_question is true when the eligibility status
% can be determined from provided facts.
can_answer_married_couples_allowance_question :-
    claims_marriage_allowance(_),
    born_before_6_april_1935(_).

% The user_is_claiming_marriage_allowance rule is true when the stored fact indicates
% that either the user or their partner currently claims Marriage Allowance 
user_is_claiming_marriage_allowance :-
    claims_marriage_allowance(true).

% The user_is_not_claiming_marriage_allowance rule is true when the stored fact indicates
% that neither the user nor their partner claims Marriage Allowance
user_is_not_claiming_marriage_allowance :-
    claims_marriage_allowance(false).


% The born_before_6_april_1935 rule is true when the stored fact
% indicates that either the user or their partner was born before 6 April 1935
born_before_6_april_1935 :-
    born_before_6_april_1935(true).

% The not_born_before_6_april_1935 rule is true when the stored
% fact indicates that neither the user nor their partner was born before 6 April 1935
not_born_before_6_april_1935 :-
    born_before_6_april_1935(false).
\end{minted}

\section{LLM Prompts}
\label{apd:prompts}
This section presents the LLM prompts used in the approaches presented above.

\begin{tcolorbox}[
    colback=gray!5,
    colframe=gray!40,
    coltitle=black,
    colbacktitle=gray!20,
    title=\textbf{NeSy-RAG Prolog Generation Prompt},
    fonttitle=\bfseries,
    breakable
]

You are an expert in Prolog programming, logic-based knowledge representation, and legal decision systems. You follow all clean code principles and best practices for knowledge engineering.

Your task is to transform the following website text into a complete and executable Prolog knowledge base intended for question answering. The knowledge base should be extensive and cover all possible user situations and questions that could arise from the provided text.

\textbf{GOAL}

The knowledge base should allow users to:
\begin{itemize}
    \item Declare information about their personal situation as required by the source text to answer all possible questions. Information is declared only using yes/no facts (e.g., \texttt{is\_minor(true).\ lives\_in\_uk(false).})
    \item Query whether specific legal conditions apply to them or whether certain governmental services, rights, or obligations are available. Evaluation should be based on user-declared facts. The conditions should exactly reflect the legal rules and criteria described in the source text.
\end{itemize}

\textbf{REQUIREMENTS}
\begin{enumerate}
    \item Clearly separate \emph{user-provided facts} from \emph{derived knowledge}.
    \item All user-provided facts must:
    \begin{itemize}
        \item Be declared as dynamic predicates including a comment that describes the predicate.
        \item Be grouped at the beginning of the knowledge base.
        \item \textbf{Must be of arity 1.}
        \item User info must \textbf{only} have one \texttt{true}/\texttt{false} argument --- users must only answer with true or false to add dynamic facts.
        \item Must contain an expressive and detailed predicate name.
        \item Must only cover the minimum necessary user information for answering questions.
    \end{itemize}
    \item Encode all rules necessary to derive legal outcomes from the user facts.
    \begin{itemize}
        \item Carefully consider if one condition suffices or if multiple conditions must be combined.
        \item Make sure for all rules to add the negated scenario (e.g., \texttt{is\_required} and \texttt{is\_not\_required}).
    \end{itemize}
    \item Use clear, consistent predicate names that reflect legal meaning.
\end{enumerate}

\textbf{OUTPUT CONSTRAINTS}
\begin{itemize}
    \item Output \textbf{only} syntactically correct and complete Prolog code.
    \item Ensure there are no infinite recursions or loops.
    \item Do not include explanations, markdown, or text outside of Prolog comments.
    \item Ensure the knowledge base can be loaded and queried without modification.
\end{itemize}

\textbf{SOURCE TEXT TO CONVERT}

\texttt{\{\{snippet\}\}}

\end{tcolorbox}

\begin{tcolorbox}[
    colback=gray!5,
    colframe=gray!40,
    coltitle=black,
    colbacktitle=gray!20,
    title=\textbf{NeSy-RAG - 0-arity Rule Generation Prompt},
    fonttitle=\bfseries,
    breakable
]

You are an expert in Prolog programming and knowledge engineering.
You receive a Prolog knowledge base and a matching text snippet of domain knowledge. Your task is to make rules that are not already 0-arity available as 0-arity rules.

To this end, your task is to generate \textbf{0-arity} Prolog rules that work as an interface to non-0-arity rules that exist in the knowledge base. Your new rules should be usable as part of a question answering system to answer yes/no questions.

Add a description to each rule that describes the content of the rule as well as the conditions under which it is fulfilled. Do not copy already existing 0-arity rules, expect them to be available to the user alongside the rules you generate.

Note that the question answering system infers an answer based on the dynamic predicates that the user answers with either \texttt{true} or \texttt{false}. Your rules must evaluate correctly according to the information the user provides.

Only answer with the rules and nothing else!

\medskip
\textbf{Text snippet from domain knowledge:}

\texttt{\{\{snippet\}\}}

\medskip
\textbf{Existing Prolog knowledge base:}

\texttt{\{\{kb\}\}}

\end{tcolorbox}

\begin{tcolorbox}[
    colback=gray!5,
    colframe=gray!40,
    coltitle=black,
    colbacktitle=gray!20,
    title=\textbf{NeSy-RAG - Query Generation Prompt},
    fonttitle=\bfseries,
    breakable
]

You are an expert in Prolog programming and knowledge engineering. Your task is to generate a valid Prolog query using the following available rules that matches the provided question of a user. Answer only with the query and nothing else. Always include the module names for the predicates in your query. Make sure that you only use the available 0-arity predicates in your query.

\medskip
\textbf{Here are the available rules:}

\texttt{\{\{zero\_arity\_rules\}\}}

\medskip
\textbf{Here is the user question you should formulate using the available rules:}

\texttt{\{\{question\}\}}

\end{tcolorbox}

\begin{tcolorbox}[
    colback=gray!5,
    colframe=gray!40,
    coltitle=black,
    colbacktitle=gray!20,
    title=\textbf{NeSy-RAG Evaluation - Predicate Answering Prompt (Simulating User)},
    fonttitle=\bfseries,
    breakable
]

Your task is to answer a yes/no dynamic predicate about a user based on the given user context.

\medskip
\textbf{Snippet:}

\texttt{\{\{snippet\}\}}

\medskip
\textbf{Initial user question:}

\texttt{\{\{question\}\}}

\medskip
\textbf{User context:}

\texttt{\{\{user\_context\}\}}

\medskip
\textbf{All available dynamic predicates:}

\texttt{\{\{dynamic\_predicates\}\}}

\medskip
Answer the following predicate as \texttt{true}/\texttt{false}, or return \texttt{None} if the information is not stated (unknown).

\medskip
\textbf{Yes/no dynamic predicate to answer:}

\texttt{\{\{dynamic\_predicate\}\}}

\end{tcolorbox}

\begin{tcolorbox}[
    colback=gray!5,
    colframe=gray!40,
    coltitle=black,
    colbacktitle=gray!20,
    title=\textbf{LLM Baseline (Upper + RAG) - Question Answering Prompt},
    fonttitle=\bfseries,
    breakable
]

You receive a text snippet, a user question, and information about the user. Your task is to answer the question if possible.

\begin{itemize}
    \item If additional information from the user that asked the question is necessary, answer with \texttt{`more'}.
    \item If the question cannot be answered because the text snippet does not contain relevant information, answer with \texttt{`None'}.
    \item If you can provide an answer, respond either with \texttt{`true'}, meaning the user question can be answered with ``yes'', or with \texttt{`false'}, meaning the user question can be answered with ``no''.
\end{itemize}

\medskip
\textbf{Here is the text snippet:}

\texttt{\{\{snippet\}\}}

\medskip
\textbf{Here is the information from the user:}

\texttt{\{\{user\_context\}\}}

\medskip
\textbf{Here is the user question you should answer:}

\texttt{\{\{question\}\}}

\end{tcolorbox}

\textbf{Note:} \texttt{user\_context} is \emph{complete} in the LLM Upper Baseline and \emph{initially empty} in the LLM RAG Baseline. The LLM RAG Baseline receives the answers to its follow-up questions as \texttt{user\_context} in the second pass.

\begin{tcolorbox}[
    colback=gray!5,
    colframe=gray!40,
    coltitle=black,
    colbacktitle=gray!20,
    title=\textbf{LLM RAG Baseline - Follow-Up Question Generation Prompt},
    fonttitle=\bfseries,
    breakable
]

You receive a text snippet, a user question, and context information from the user. Your task is to generate a list of yes/no follow-up questions that the user should answer such that we can answer their initial question.

\medskip
\textbf{Text snippet:}

\texttt{\{\{snippet\}\}}

\medskip
\textbf{Initial user question:}

\texttt{\{\{question\}\}}

\medskip
\textbf{Known information from the user:}

No user context acquired yet.

\end{tcolorbox}

\begin{tcolorbox}[
    colback=gray!5,
    colframe=gray!40,
    coltitle=black,
    colbacktitle=gray!20,
    title=\textbf{LLM RAG Baseline Evaluation - Follow-up Question Answering Prompt},
    fonttitle=\bfseries,
    breakable
]

You receive a question and information about the user. Your task is to answer the question based on the information that is provided about the user.

\medskip
\textbf{Question:}

\texttt{\{\{followup\_question\}\}}

\medskip
\textbf{User information:}

\texttt{\{\{user\_context\}\}}

\end{tcolorbox}

\begin{tcolorbox}[
    colback=gray!5,
    colframe=gray!40,
    coltitle=black,
    colbacktitle=gray!20,
    title=\textbf{Retrieval Snippet Relevance Classification Prompt},
    fonttitle=\bfseries,
    breakable
]

You receive a question and a text snippet. Your task is to decide if the text snippet contains the rules for answering the question. Answer with \texttt{`true'} if the snippet is relevant and answer with \texttt{`false'} if the snippet is not relevant.

\medskip
\textbf{The snippet:}

\texttt{\{\{snippet\}\}}

\medskip
\textbf{The question:}

\texttt{\{\{question\}\}}

\end{tcolorbox}

\end{document}